\documentclass[a4paper,conference]{IEEEtran}
\usepackage{ifpdf}

\usepackage{cite}

\ifCLASSINFOpdf
   \usepackage[pdftex]{graphicx}
   \DeclareGraphicsExtensions{.pdf,.jpeg,.png}
\else
    \usepackage[dvips]{graphicx}
   \DeclareGraphicsExtensions{.eps}
\fi
\usepackage{xcolor}
\usepackage{bbm}
\usepackage{amsmath}
\usepackage{algorithmic}

\usepackage{array}

\ifCLASSOPTIONcompsoc
 \usepackage[caption=false,font=normalsize,labelfont=sf,textfont=sf]{subfig}
\else
 \usepackage[caption=false,font=footnotesize]{subfig}
\fi
\usepackage{fixltx2e}
\usepackage{dblfloatfix}

\usepackage{url}

\begin{document}
%

\title{Privileged Attribution Constrained Deep Networks for Facial Expression Recognition}

\author{\IEEEauthorblockN{Jules Bonnard\IEEEauthorrefmark{1},
Arnaud Dapogny\IEEEauthorrefmark{2},
Ferdinand Dhombres\IEEEauthorrefmark{3},
Kevin Bailly\IEEEauthorrefmark{1}\IEEEauthorrefmark{2}}
\IEEEauthorblockA{\IEEEauthorrefmark{1}Sorbonne Universite, CNRS, ISIR, F-75005 Paris, France\\ Email: \{jules.bonnard, kevin.bailly\}@sorbonne-universite.fr}
\IEEEauthorblockA{\IEEEauthorrefmark{2}Datakalab, F-75017 Paris, France\\
Email: arnaud.dapogny@gmail.com}
\IEEEauthorblockA{\IEEEauthorrefmark{3}Sorbonne Université, INSERM U1142 LIMICS, GRC26,  F-75005 Paris, France\\
Email: ferdinand.dhombres@inserm.fr}
}


\maketitle

\begin{abstract}
Facial Expression Recognition (FER) is crucial in many research domains because it enables machines to better understand human behaviours. 
FER methods face the problems of relatively small datasets and noisy data that don't allow classical networks to generalize well. To alleviate these issues, we guide the model to concentrate on specific facial areas like the eyes, the mouth or the eyebrows, which we argue are decisive to recognise facial expressions.
We propose the Privileged Attribution Loss (PAL), a method that directs the focus of the model towards the most salient facial regions by encouraging its attribution maps to correspond to a heatmap formed by facial landmarks. 
Furthermore, we introduce several channel strategies that allow the model to have more degrees of freedom. The proposed method is independent of the backbone architecture and doesn't need additional semantic information at test time. Finally, experimental results show that the proposed PAL method outperforms current state-of-the-art methods on both RAF-DB and AffectNet.
\end{abstract}


%
\IEEEpeerreviewmaketitle

\section{Introduction}

Since 2012 and Alex Krizhevsky et al. \cite{krizhevsky_imagenet_2012} won the
\textit{ImageNet Challenge}, Deep Convolutional Neural Networks (CNN) have obtained state-of-the-art results in almost all computer vision tasks thanks to large amounts of data. 
However, for tasks like Facial Expression Recognition (FER), which have comparatively smaller available datasets, CNNs have trouble reaching the levels of performance achieved in other domains such as visual object recognition. CNNs also have trouble dealing with noisy data, which is common in FER datasets, where some facial areas can be occluded.  But this structure can specifically be taken profit of to help explain and improve the results of a CNN on FER, by guiding the learning mechanism towards certain discriminative areas of the input images.\par
\begin{figure}[!t]
\centering
\includegraphics[width=3.4in]{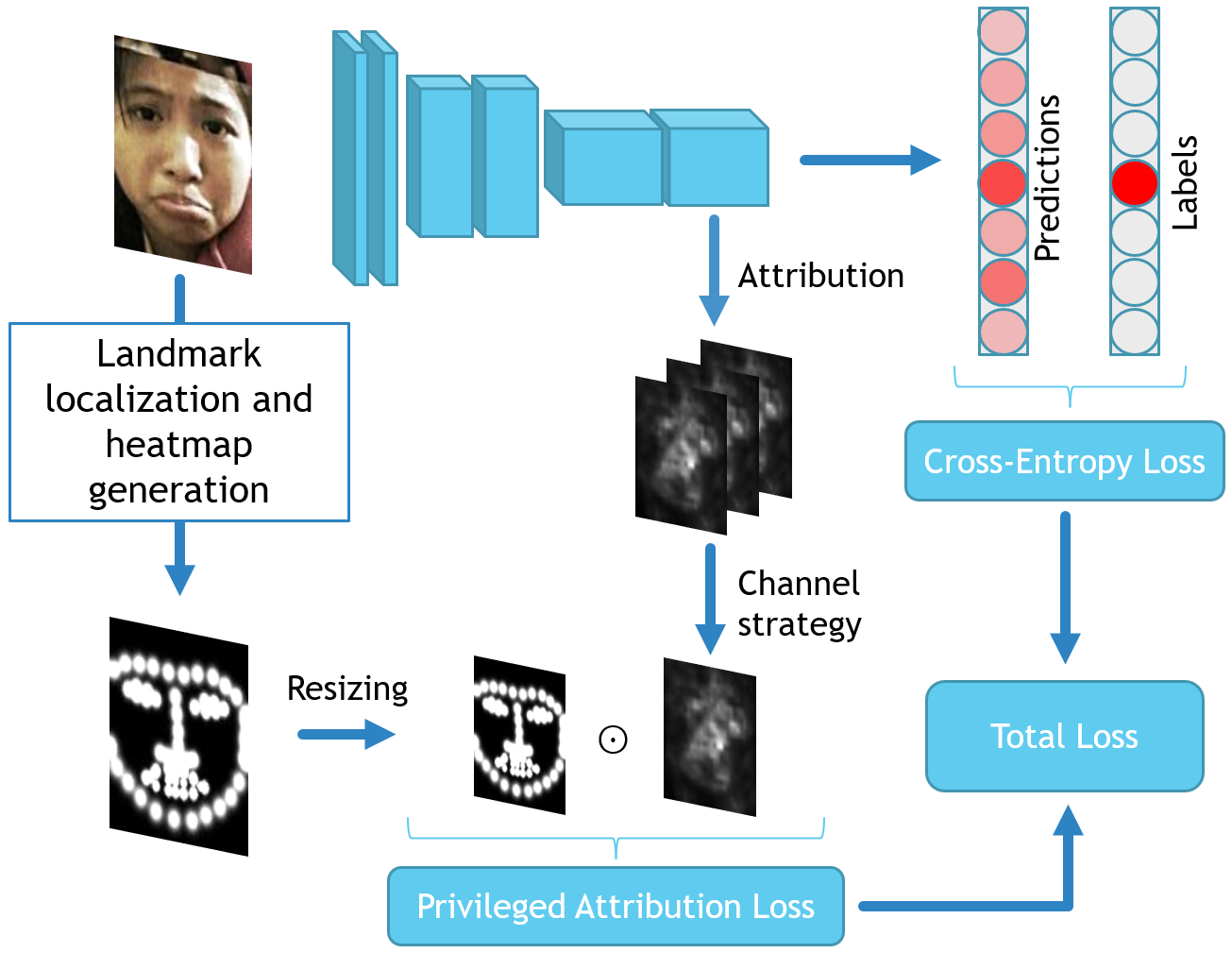}
\caption{Overview of the proposed PAL method. At train time, the input image is passed through a CNN and the output is given to a classic cross-entropy loss. The attribution map of a certain feature map w.r.t. the output is then computed. The cross-correlation between an attribution map, chosen through a channel strategy, and an external facial landmark heatmap, is then calculated.}
\label{schema}
\end{figure}

We use the fact that these deep networks can be written as a composition of functions that output \textit{feature maps}, which are essentially shaped like images, with several channels. These feature maps can be used to understand which pixels have the most influence on the final predictions by applying several different gradient-based methods \cite{ancona2018better} on them. We can then create what are called \textit{attribution maps}, which offer the influence of a certain feature map on the model's predictions. In this paper, we introduce a novel loss that we call the "Privileged Attribution Loss" (PAL) that forces these attribution maps to resemble a certain given prior, and therefore forces the whole model to focus more on certain pixels. We apply PAL to FER by using a heatmap of facial landmarks as a semantic prior. The method therefore leverages spatial information during the training to predict facial expressions, without needing this prior at test time.
Figure \ref{schema} summarizes the proposed method.\par

In summary, the paper proposes the following contributions:
\begin{itemize}
    \item We introduce PAL, a loss that consists in maximising the cross-correlation between different attribution maps and an external heatmap shaped prior.
    \item We propose different channel strategies that allow the model to have more liberty whilst improving its predictive performances.
    \item Experimentally, we then apply PAL to FER. We evaluate the influence of the choice of the layer and the channels to which we apply the PAL. We also evaluate the choice of the attribution method. Results that consistently outperform our baselines and extend the state-of-the-art on important FER datasets such as RAF-DB and AffectNet verify the interest of our method.
\end{itemize}

The paper is organized as follows: first of all, we will review related works on the subject of spatially constrained methods. Secondly, we will present the inner workings of our method, and finally, we will demonstrate our experimental results with a qualitative and quantitative analysis.\par

\section{Related Work}
In this work, we aim at spatially constraining a FER model by guiding its attribution maps. Thus, we will first present methods that spatially constrain their models for Facial Expression tasks, then we will focus on methods that explicitly constrain their attributions.

\subsubsection{Constrained Learning for Facial Expression tasks}

Spatially-constrained methods have become very popular for FER tasks as the spatial repartition of features is paramount \cite{dapogny2018confidence}. First, some methods rely on guiding their model with explicit spatial modules: Zhao \textit{et al.} \cite{zhao2021robust} propose to dynamically learn to focus on local and global salient facial features with a local-feature extractor.
These methods learn the important spatial features without any semantic prior, and therefore don't guarantee the relevance of the salient areas w.r.t. the task at hand. Second, some methods use semantic information learnt from auxiliary tasks to improve their predictions w.r.t. another, main task: Shao \textit{et al.} proposed JAA-Net \cite{jaanet}, which aims to predict Facial Action Units by first learning to align facial landmarks. They then use the extracted features to refine attention maps and guide the facial action unit detection.
Pu \textit{et al.} \cite{pu2021auexpression} have proposed a FER method that constrains the model by predicting facial action units first. However, there is a risk of propagating errors between the action unit and FER prediction modules.\par
Another way of guaranteeing that the model focuses on the most salient areas can be to directly take advantage of additional information instead of trying to model it with an intermediate network. Jacob \textit{et al.} \cite{Jacob_2021_CVPR} employed heatmaps in which the areas relative to specific facial action units are highlighted, as a prior information for facial action units detection. In this method, they constrain the attention maps produced by a Transformer encoder to resemble the prior information heatmaps. This allows the model to focus on specific areas that are key to predict the presence of an action unit using an external information, without needing this information at inference time.
These heatmaps are obtained using an attention mechanism that is used to select the most relevant parts of an input based on another (cross-attention) or the same input (self-attention). However, this doesn't guarantee that the model directly focuses on these salient areas.
On the contrary, PAL leverages additional semantic information in a more straightforward way by guiding its attribution maps, and therefore the exact areas that the model concentrates on. Therefore, in the next section, we will present methods that constrain their attribution maps to guide their learning.\par


\subsubsection{Constrained Attribution Learning}

Attribution methods
aim at evaluating the impact of each input feature on the predictions. These methods have originally been used to explain network predictions, but have also recently been used to constrain the learning of these models. These models either use occlusion-based or gradient-based attribution methods.
Occlusion-based methods \cite{zeilerocclusion} evaluate the attribution of a certain patch of pixels by comparing the predictions given by the input image with the predictions given by the occluded input image. Du \textit{et al.} \cite{du2019learning} applied this natural language processing by reducing the score of non-important features using expert-annotated important clauses in the input sentences. However, occlusion-based attribution methods are usually iterative and involve prohibitive computational costs especially on images.\par
To reduce the runtime, gradient-based attribution methods were introduced. Simonyan \textit{et al.} \cite{simonyan2014deep} first introduced image-specific class saliency maps. Selvaraju \textit{et al.} \cite{grad_cam} proposed \textit{Grad-CAM}, a method that consists in computing class-specific heatmaps by multiplying each channel of the last convolutional feature map by its "importance" (the mean over that specific channel of the gradient of the chosen output class w.r.t. the last convolutional layer's output). In 2016, Shrikumar \textit{et al.} \cite{shrikumar2017just} proposed a technique to sharpen the attribution maps called \textit{Gradient*Input}, where the partial derivatives of the output are multiplied by the input features. The idea is that the gradient tells us how important a certain feature is, and the input tells us how strongly it is expressed in the final prediction. 
Other methods have been introduced more recently like \textit{Integrated Gradients} (Sundarajan \textit{et al.} \cite{sundararajan2017axiomatic}), which is more stable than \textit{Gradient*Input} and generally better captures the explanation of each input feature, at the cost of a larger computational load. \par



For efficiency reasons, most approaches nowadays use gradient-based attribution methods instead of occlusion-based methods for attribution constraining. Nevertheless, there are different ways to constrain this attribution. Erion \textit{et al.} \cite{erion2020improving} decided to constrain their model so that its attributions would satisfy desirable properties, \textit{i.e.} sparsity and smoothness. However, they do not use extra information which could help them in identifying key features. By contrast, Liu \textit{et al.} \cite{liu2019incorporating} applied attribution constraining to NLP and added a L2 norm between their attribution (Integrated Gradients) and an external prior information (toxic or non-toxic words). In the same vein, Ross \textit{et al.} \cite{ross2017right} introduced the idea of being "right for the right reasons" (RRR) for Computer Vision. They proposed to regularize the gradients w.r.t. the input features by penalizing the 'unimportant' features with a prior binary mask. This mask indicates whether or not a certain feature should be relevant for a certain prediction. However, in their paper, they only constrain their gradient inputs to improve their explanations, and not to improve the performance.
By contrast, in this work, we use the attribution maps to spatially constrain the learning process in order to improve the predictive power of a FER model. In the next section, we introduce our methodology, first by focusing on the different possible attribution methods, then our loss and channel strategy. Finally, we will explain our heatmap prior information.

\noindent
\begin{figure}[!t]
\centering
\subfloat{\includegraphics[width=0.44in]{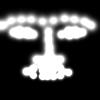}%
\label{h1}}
\hfil
\subfloat{\includegraphics[width=0.44in]{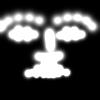}%
\label{h2}}
\hfil
\subfloat{\includegraphics[width=0.44in]{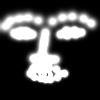}%
\label{h3}}
\hfil
\subfloat{\includegraphics[width=0.44in]{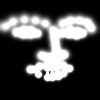}%
\label{h4}}
\hfil
\subfloat{\includegraphics[width=0.44in]{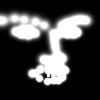}%
\label{h5}}
\hfil
\subfloat{\includegraphics[width=0.44in]{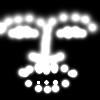}%
\label{h6}}
\hfil
\subfloat{\includegraphics[width=0.44in]{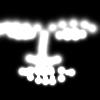}%
}
\hfil
\subfloat{\includegraphics[width=0.44in]{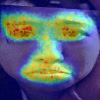}%
\label{h7}}
\hfil
\subfloat{\includegraphics[width=0.44in]{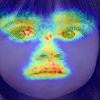}%
\label{h8}}
\hfil
\subfloat{\includegraphics[width=0.44in]{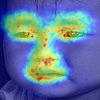}%
\label{h9}}
\hfil
\subfloat{\includegraphics[width=0.44in]{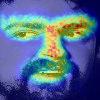}%
\label{h10}}
\hfil
\subfloat{\includegraphics[width=0.44in]{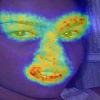}%
\label{h11}}
\hfil
\subfloat{\includegraphics[width=0.44in]{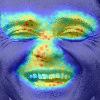}%
\label{h12}}
\hfil
\subfloat{\includegraphics[width=0.44in]{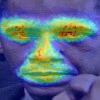}%
\label{h13}}
\caption{Examples of heatmap priors of facial landmarks (top) and attribution maps on RAF-DB (bottom) for a model trained with PAL. By constraining the attribution maps with an external information (in our case, the localization of facial landmarks), we explicitly guide the network towards using more prominent facial regions (e.g. mouth or eye regions), significantly improving FER without requiring additional information at inference time.}
\label{heatmaps}
\end{figure}

\section{Method}
\subsection{Attribution Methods}

All Convolutional Neural Networks can be written as a composition of functions \cite{chen:hal-02507512}. Let $f$ denote a CNN, then $f$ can be re-written as follows:
\begin{equation}
f(I) = f^L \circ f^{L-1} \circ ... \circ f^1(I)
\end{equation}
\noindent
where $I$ denotes the input image, $L$ denotes the total number of functions and $f^l$ denotes a function, usually a convolutional layer, an activation function, a pooling function or a batch normalization. $\Theta$ denotes the set of trainable parameters of the whole network $f$. With this in mind, we can extract feature maps after the activation functions which are essentially built like images (height*width*channels). These feature maps can be analysed, and their \textit{contribution} or \textit{attribution} can be computed with several attribution methods. These methods are able to evaluate the relevance of certain pixels of the input image, but they can also be used to evaluate the relevance of the pixels of the intermediate feature maps.
In this work, we choose to use these attribution scores to guide the learning, and not only for model evaluation purposes. We will therefore introduce the different attribution methods used in this paper.\par
The first and more straightforward of these approaches (which will be denoted as \textit{Grad} in what follows) consists in computing the absolute value of the gradient of the sum of all coordinates $f_o$ of the output vector $f(I)$ w.r.t. each pixel of $f^l$. \linebreak
We choose to compute the derivative of $\Sigma f_o$ to guide all of the network's predictions. Let $a_{i,j,c}^l$ denote the attribution of the pixel $(i,j)$ of the $c$-th channel of the feature map given by layer $l$. The \textit{Grad} attribution method can be written as:

\begin{equation}
a_{i,j,c}^l(I) = |\frac{\partial \Sigma f_o}{\partial f_{i,j,c}^l}(I)|
\end{equation}

As shown in \cite{chen:hal-02507512}, any ReLU-based CNN can be decomposed as a piece-wise affine function of each pixel of $f^l$ and the contribution of each pixel is exactly the following:
\begin{equation}
a_{i,j,c}^l(I) = |\frac{\partial \Sigma f_o}{\partial f_{i,j,c}^l}(I)| \cdot f_{i,j,c}^l(I)
\end{equation}
\noindent
This will be denoted as the \textit{Grad*Input} model.\par
One intuition of the difference between the two is that \textit{Grad} corresponds to how a small change in the input will impact the output of the network, whereas \textit{Grad*Input} corresponds to the total contribution of a feature on the output of the network.\par 

\subsection{Privileged Attribution Loss}

Let $a^*$ denote a standardized prior map on the attribution model. The loss we build aims to make the model "pay more attention" to certain regions in the input image, those regions being the ones highlighted in $a^*$. To incentivize that behaviour, we maximize the cross-correlation between the attribution maps $a^l$ and the prior maps $a^*$. If we have $\mu (a^l) = \Sigma_{i,j} a^l_{i,j,c}$ and $\sigma^2(a^l) = \sum_{i,j}(a^l_{i,j,c} - \mu (a^l))(a^l_{i,j,c} - \mu (a^l))$, then:

\begin{equation}
\Lambda_{PAL}^l(\Theta) = -\sum_{i,j,c}\frac{a^l_{i,j,c} - \mu(a^l)}{\sigma(a^l)} * a^*_{i,j,c}
\end{equation}

The rationale behind using a cross-correlation term instead of e.g. a L2 norm matching \cite{liu2019incorporating} is to provide an additional degree of freedom to the network, as we want the constrained attribution map to have the same profile as the prior, regardless of its dynamics. The \textit{PAL} term of the loss gets added to a cross-entropy term for classification.

\noindent

\subsection{Channel Strategies}

In what precedes we have constrained the attribution $a^l_{i,j,c}$ of each pixel $(i,j)$ of each channel $c$ to correspond to a certain prior $a^*$. However, forcing all channels to be similar to a certain prior $a^*$ (we will call that the \textit{All Channels} strategy) might be too strong a constraint. A weaker formulation consists in forcing the \textit{mean} of the attributions across all channels to resemble a prior $a^*$. We will call that the \textit{Mean} strategy. In this case, with $C$ being the number of channels of $f^l$, the attribution becomes:

\begin{equation}
a_{i,j}^l = \frac{1}{C} \sum_{c=1}^C a_{i,j,c}^l 
\end{equation}

This again might be too strong a constraint, and we could only force a fraction of the channels to resemble $a^*$. This would allow certain channels to stay free and learn other patterns that might be important for the predictions. The attribution model would then look like this, with $C_1<C$:

\begin{equation}
a_{i,j}^l = \frac{1}{C_1} \sum_{c=1}^{C_1} a_{i,j,c}^l
\end{equation}
\noindent
In what follows, we set $C_1 = \frac{C}{2}$ and refer to this approach as the \textit{Mean of half} strategy.

\subsection{Prior heatmap}

The method presented in this paper is generic and can be applied to many computer vision tasks. Its strength comes from the fact that it uses prior information of particularly relevant regions at train time to constrain the attribution of feature maps in the network, while no additional information is needed at inference time. The choice of the prior might orientate the final decisions a lot.\par
In this work, we choose to use face alignment points as a prior. 
From a 2D face image \textit{I}, the face alignment model \textit{g} locates 68 facial landmarks that form the face shape $\textbf{y} \in \mathbf{R}^{68^*2}$ (the \textit{n}-th row of this matrix corresponding to the 2D coordinates of the \textit{n}-th facial landmark).\newline
Let $\mathbf{1}^L_{i,j}$ denote the indicator function which equals 1 when on a landmark, and 0 otherwise.
From these facial landmark coordinates, we create an image $a^*$ with:
\begin{equation}
    a^*_{i,j}= \mathbf{1}^L_{i,j}
\end{equation}
\noindent
Which ultimately gives an image with pixel values of 1 at the landmarks and 0 elsewhere.
We then apply a gaussian filter with a standard deviation $\sigma$ of 3 to this image $a^*$. We then have:

\begin{equation}
a^{*filtered}_{i,j} = \sum_{k=1}^{68} \frac{1}{\sqrt{2\pi\sigma^2}}exp(-\frac{(i-y_{k,1})^2+(j-y_{k,2})^2}{2\sigma^2})
\end{equation}

Examples of such prior heatmaps are shown in Figure \ref{heatmaps}. Notice how the network manages to integrate the prior information, and focuses on the highlighted relevant localisations. As we will show in the experiments, this allows to dramatically improve the FER performance, without needing additional information at inference time.

\section{Experiments}
In this section, we validate our method on two of the most recent and challenging datasets for FER. We also conduct an ablation study to validate the parameters of PAL
such as the attribution method, the layer to which the method is applied and the channel strategy.

\begin{figure}[!t]
\centering
\includegraphics[width=2.8in]{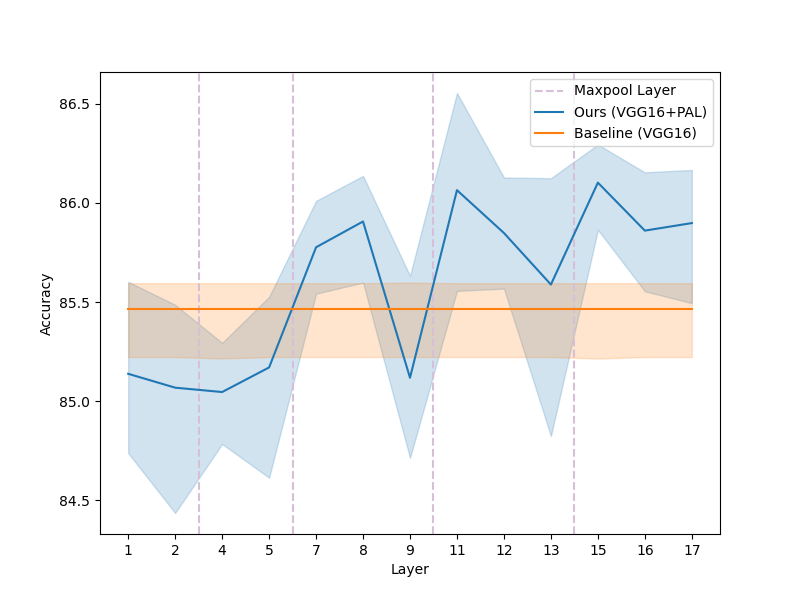}
\caption{Performance of the model learnt with PAL applied to different convolutional layers of the VGG with \textit{Grad} attribution method and \textit{Mean of Half} channel strategy. The plot shows the mean accuracy score and 95\% confidence interval.}
\label{layers}
\end{figure}

\begin{table}[!t]
    \renewcommand{\arraystretch}{1}
	\caption{Ablation study on RAF-DB dataset comparing different attribution methods and channel strategies.}
	\label{ablation}
	\centering
	\resizebox{8.2cm}{!}{
    \begin{tabular}{|c||c|c|c|c|c||c||c|}
    \hline
    \textbf{Method} & $Attribution$ & $Channels$ & $Acc$ \\
    \hline
    $VGG16 $ & $---$ & $---$ & $85.4 \pm 0.2$ \\
    \hline
    $VGG16  + PAL$ & $Grad$ & All Channels & $85.31 \pm 0.23$ \\
    $VGG16  + PAL$ & $Grad$ & Mean & $86.21 \pm 0.12$ \\
    $VGG16  + PAL$ & $Grad*Input$ & Mean & $ 86.38 \pm 0.17 $\\
    $VGG16  + PAL$ & $Grad$ & Mean of half & $86.46 \pm 0.13$ \\
    $VGG16  + PAL$ & $Grad*Input$ & Mean of half & $86.82 \pm 0.1$ \\
    
    \hline
    \end{tabular}}

\end{table}

\subsection{Databases}
\subsubsection{RAF-DB}
Real-world Affective Faces Database (RAF-DB) is a large-scale facial expression database. Unlike most other FER datasets, all of its images have been manually annotated by several annotators. 
It is annotated with seven different emotions (sadness, angriness, happiness, surprise, disgust, fear, neutral). It contains a train set of 12271 aligned images, and a test set of 3068 images with the same label distribution as the train set.\par

\subsubsection{AffectNet}

AffectNet \cite{affectnet} is the largest FER database with 400k face images collected through webscrapping techniques. These have been manually annotated by 12 annotators for the 7 basic emotions. We only use the images annotated for the 7 basic emotions, which represent 280k training samples and 3.5k test samples.

\subsection{Implementation Details}
For RAF-DB, we train our model using the ADAM optimizer with a batch size 16 and a base learning rate 5e-5 with polynomial decay. Face images are resized to 224x224, and augmented with random rotation [-10$^{\circ}$, 10$^{\circ}$] followed by a random horizontal flip. The best model was chosen after 75 epochs by keeping the weights that maximized the accuracy on a validation set sampled with the same label distribution as the test set. For AffectNet we used the same parameters, we used a base learning rate of 4e-5 and kept the best model after 40k iterations with class-balanced mini-batches.
The ablation study was conducted using a VGG16 backbone pre-trained on VGGFace \cite{vggface}. To improve the results and approach those obtained by the state-of-the-art methods, we employed a ResNet50 backbone pre-trained on VGGFace \cite{vggface}. For RAF-DB, we use an off-the-shelf alignment method \cite{arnaud2019treegated} to fetch facial landmarks and create heatmaps from them. The face alignment points are made available for AffectNet.



\subsection{Ablation Study}
\subsubsection{Impact of the layer to which PAL is applied}

First, we look at the impact of the choice of the layer on which we apply PAL.
We observe on Figure \ref{layers} an important performance boost when the method is applied on the last layers. 

This could be explained by the fact that the information in the early layers is very low-level, and therefore the model won't be able to focus on the important areas. Selvaraju \textit{et al.} \cite{grad_cam} point out that the last convolutional layers of a CNN are a good "compromise between high-level semantics and detailed spatial information". Another reason is that in the first layers there are very few channels, so even using the "Mean of Half" channel strategy might be too strong a constraint.\par
Finally, we observe a drop in performance for layers 9 and 13, which can be explained by the fact that these layers precede a max-pooling layer, represented in Figure \ref{layers} by vertical lines. This means that their attribution maps are sparse (three pixels out of four have a zero value for a $(2, 2)$ pooling layer) and therefore can't take the appearance of our semantic prior. For layers 2 and 5 that also come before a pooling layer, we observe that the variance of these values is higher.\par

\subsubsection{Attribution Methods}

To explore the impact of using different attribution methods, we train our model with \textit{Grad} and \textit{Grad*Input}. We see on Table \ref{ablation} that \textit{Grad*Input} works better whichever channel strategy we use. This could be explained by the fact that the former outputs sharper attribution maps (\cite{ancona2018better}, \cite{shrikumar2017just}). The gradient accounts for the importance of a certain feature in the output prediction, and the input accounts for how strongly it is expressed in the output prediction.

\subsubsection{Channel Strategy}

\begin{table}[!t]
    \renewcommand{\arraystretch}{1}
	\caption{Ablation study on RAF-DB dataset comparing of different values for $C_1$ on the channel strategy.}
	\label{ablation_channels}
	\centering
	\resizebox{5.8cm}{!}{
    \begin{tabular}{|c||c|c|c|c|c||c||c|}
    \hline
    \textbf{Method} & $C_1$ & $Acc$ \\
    \hline
    $VGG16 $ & $---$ & $85.4 \pm 0.2$ \\
    \hline
    $VGG16  + PAL$ & $C/4$ & $86.55 \pm 0.56$ \\
    $VGG16  + PAL$ & $C/2$ & $86.82 \pm 0.1$ \\
    $VGG16  + PAL$ & $3C/4$ & $ 86.32 \pm 0.25 $\\
    $VGG16  + PAL$ & $C$ & $ 86.38 \pm 0.17 $\\
    
    \hline
    \end{tabular}}
\end{table}

We also evaluate the impact of different channel strategies and the choice of $C_1$ on our method. We can see on Table \ref{ablation} that forcing all channels to resemble our prior information heatmap is too strong a constraint. However, Table \ref{ablation_channels} shows that taking $C_1=C$ (\textit{Mean} strategy) improves the baseline accuracy by \text{0.98\%}, which indicates that a weaker constraint also improves the predictive power of the model. We also decide to evaluate the impact of different values of $C_1$. We see that taking $C_1=3C/4$ gives similar results to the \textit{Mean} strategy, which suggests that the constraint on the channels is still too strong. However, taking $C_1=C/4$ improves the baseline accuracy by $1.15\%$ which shows that the freedom given to some of the channels can be beneficial for the model. Finally, the best results are obtained with $C_1=C/2$ (\textit{Mean of Half} strategy) which improves the baseline accuracy score by \text{1.42\%}.
Ultimately, we see that the \textit{Mean of Half} strategy significantly outperforms the baseline thanks to a good compromise between guiding the model towards salient facial areas and letting it explore other features.

\subsubsection{Sensitivity to landmarks}

\begin{figure}
\centering
\includegraphics[width=2.8in]{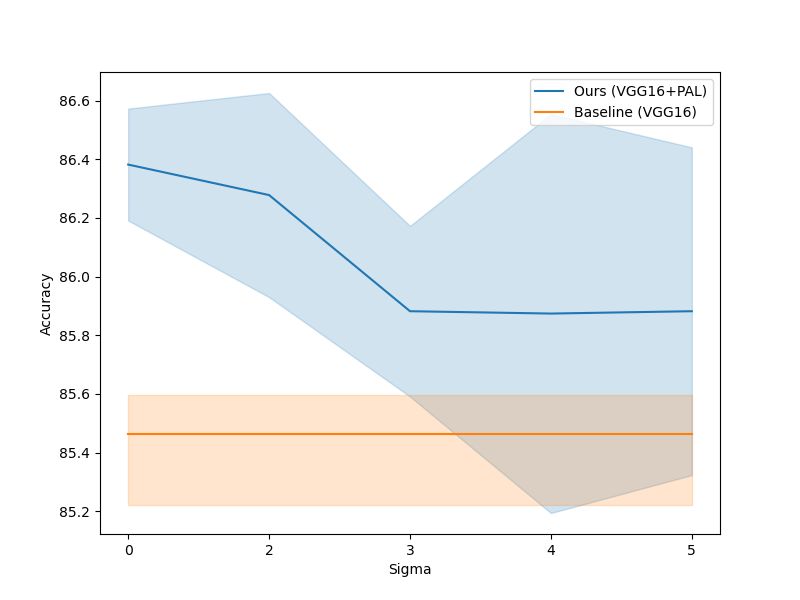}
\caption{Performance of the model learnt with heatmaps created with noisy landmarks (evaluation for different values of sigma).}
\label{fig:sensi_landmarks}
\end{figure}

\begin{figure}[!t]
\centering
\subfloat{\includegraphics[width=0.5in]{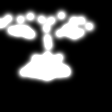}%
\label{sigma1}}
\hfil
\subfloat{\includegraphics[width=0.5in]{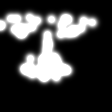}%
\label{sigma2}}
\hfil
\subfloat{\includegraphics[width=0.5in]{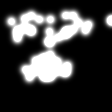}%
\label{sigma3}}
\hfil
\subfloat{\includegraphics[width=0.5in]{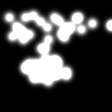}%
\label{sigma4}}
\hfil
\subfloat{\includegraphics[width=0.5in]{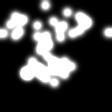}%
\label{sigma5}}
\caption{Examples of noisy heatmaps. The landmark points were sampled from a gaussian distribution with values of sigma going from 1 to 5 (from left to right).}
\label{noisy_heatmaps}
\end{figure}

\begin{table}[!t]
    \renewcommand{\arraystretch}{1.1}
	\caption{Comparison on RAF-DB and AffectNet. Accuracy scores in \%}
	\label{tab:rafdb_res}
	\centering
	\resizebox{6.2cm}{!}{
    \begin{tabular}{|c||c|c|}
    \hline
    \textbf{Method} & RAF-DB & AffectNet\\
    \hline
    IPA2LT \cite{Zeng_2018_ECCV} & 86.77 & 57.31 \\
    THIN \cite{THIN} & 87.81 & 63.97 \\
    DACL \cite{Farzaneh_2021_WACV} & 87.78 & 65.20 \\
    EfficientFace \cite{zhao2021robust} & 88.36 & 63.70 \\
    PSR \cite{psr} & 88.98 & 63.77\\
    PAENet \cite{PAENet} & -- & 65.29 \\
    DMUE \cite{DMUE} & 89.42 & --\\
    FDLR \cite{fdlr} & 89.47 & --\\
    \hline
    Baseline (Resnet50) & 88.6 & 62.14\\
    \hline
    Ours (Resnet50 + PAL) & \textbf{89.54} & \textbf{65.83} \\
    \hline
    \end{tabular}
    }
\end{table}	

\begin{figure}[!t]
\centering
\subfloat{\includegraphics[width=44mm]{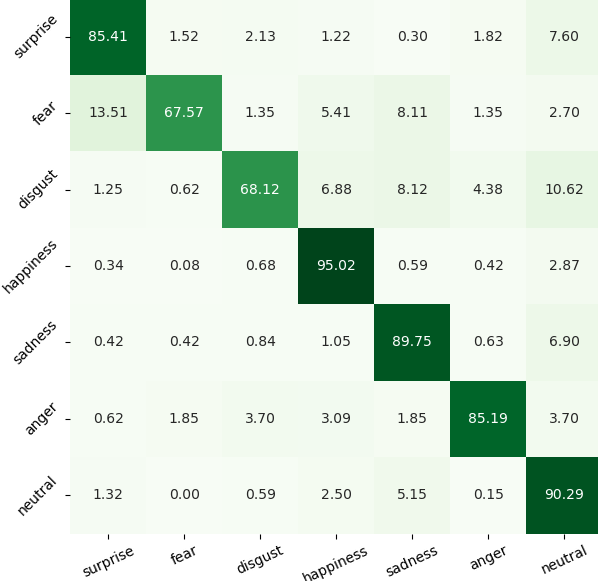}%
\label{conf_raf}}
\hfil
\subfloat{\includegraphics[width=44mm]{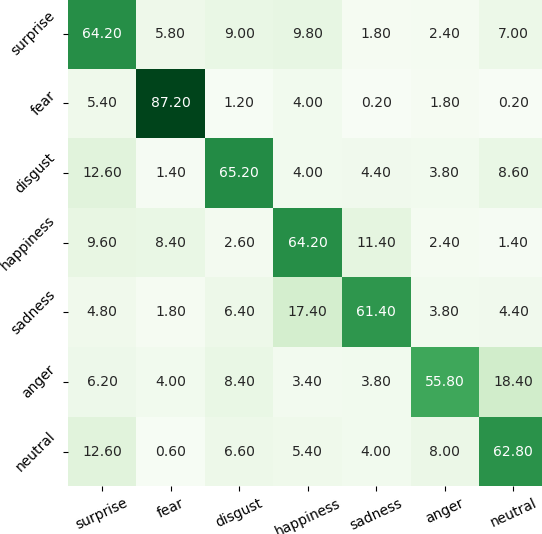}%
\label{conf_affectnet}}
\caption{Confusion matrices on the test sets of RAF-DB (left) and AffectNet (right).}
\label{confusion_matrices}
\end{figure}

\begin{figure*}[!t]
\centering     
\includegraphics[width=130mm]{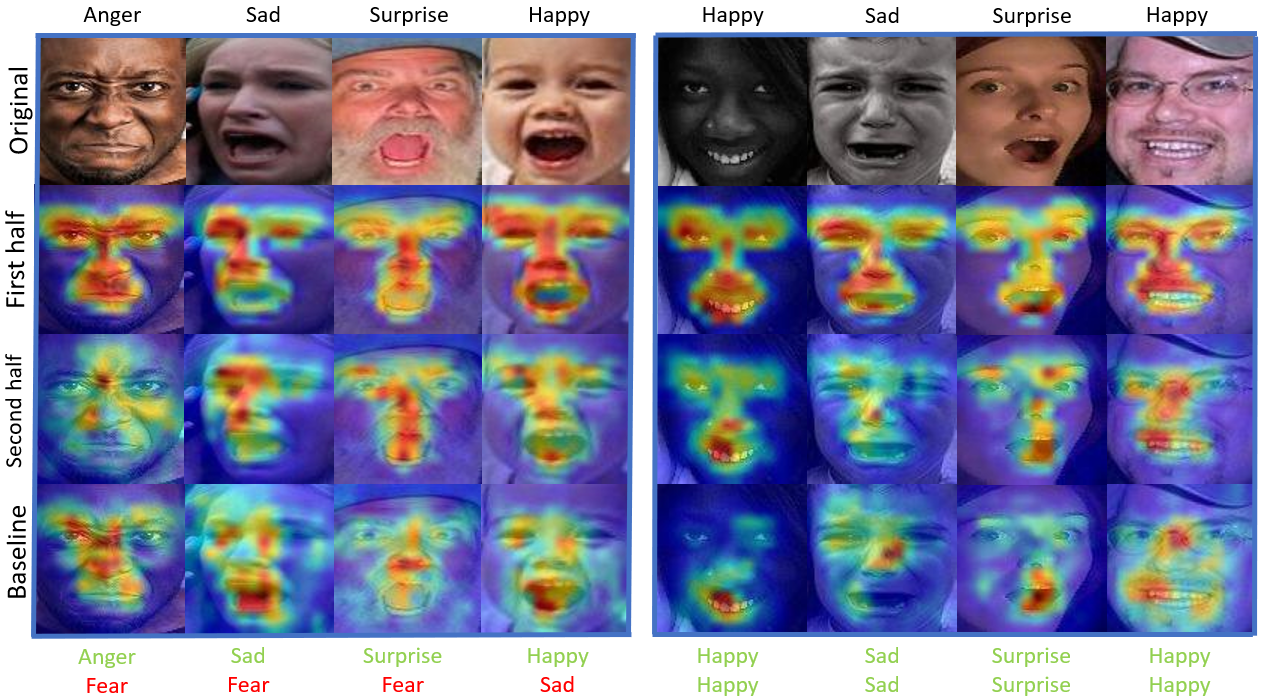}
\caption{On the first row are examples are chosen from the RAF-DB test set, with the ground truth expressions above them. On the second and third rows, examples of \textit{Grad*Input} attribution maps for the PAL-trained model with \textit{Mean of Half} channel strategy. The second row shows the mean map for the first half of the channels (constrained) while the third row shows the mean map for the second half of the channels (not constrained). On the fourth row, \textit{Grad*Input} attribution maps without PAL. Under the images are the predictions from the model trained with Pal (top) and without (bottom).}
\label{half_maps_fig}
\end{figure*}

We conducted an experiment to evaluate the sensitivity of the method to the correctness of the privileged information. To do this, we created heatmaps derived from noisy landmarks (we sampled from a gaussian distribution with sigma equal to 2, 3, 4 and 5). Examples of these heatmaps can be found in Figure \ref{noisy_heatmaps}. We can observe from Figure \ref{fig:sensi_landmarks} that the model learnt with the correct heatmaps (sigma=0) performs better than with the incorrect heatmaps. Even though the mean accuracy score of this method decreases while the value of sigma increases, we see that the model performs better than the baseline, which can be explained by the fact that the heatmaps still highlight interesting areas of the face. Furthermore, this shows that the method is robust to certain inaccuracies in the prior information.

\subsection{Comparison with State-of-the-art Methods}

Table \ref{tab:rafdb_res} compares our best results with existing baselines as well as current state-of-the-art methods on the RAF-DB and AffectNet databases. First, adding PAL significantly improves the already competitive baselines on both RAF (+$0.96\%$) and AffectNet (+$3.69\%$) with the exact same number of parameters, memory footprint and inference time. Also, the privileged information (in this case, the localisation of the facial landmarks) is necessary for training with PAL but not at inference time, hence this performance upgrade comes at virtually no cost in inference.

Second, for RAF-DB, we extend the current state-of-the-art by improving the top accuracy up to $89.54\%$. On AffectNet, our method outperforms the previous best result by improving the top accuracy from $65.40\%$ to $65.83\%$. This is presumably thanks to the fact that, unlike its closest frontrunners (DMUE \cite{DMUE} and FDLR \cite{fdlr} on RAF-DB, and PAENet \cite{PAENet} on AffectNet), the proposed method forces the model to focus on the most salient facial areas \textit{a priori} by guiding its attribution, while still letting the model have some freedom in its attribution thanks to the proposed Mean of half strategy. Last but not least, note that the proposed improvement could in theory be used along with these methods, \textit{i.e.} distillation \cite{distilled} or identity-aware training \cite{THIN}.

\subsection{Qualitative Analysis}

\subsubsection{Attribution methods}

PAL aims to improve the predictions of a model by guiding its attribution maps. Figure \ref{half_maps_fig} shows that, when using PAL, the attribution maps are generally neater and the predictions are usually more accurate. Notice, for example, how, e.g. on column 4, the baseline wrongly predicts the Sad class whereas our model correctly predicts Happy. Likely, this is due to the fact that PAL allows better repartition of the features throughout the relevant face regions (mouth, eyes, and so on), which in turn improves the predictive accuracy, echoing the work of Dapogny \textit{et al.} \cite{dapogny2016confidenceweighted}. This is the case for the first half of the channels for a method trained with the proposed Mean of half strategy. Furthermore, this strategy allows to provide the model freedom to put emphasis on certain face regions when it is needed, e.g. as shown on columns 5-8, also contributing to the accuracy improvement.

\section{Conclusion}
In this paper, we presented a novel Privileged Attribution Loss for FER. PAL consists in maximizing the cross-correlation between the attribution maps of a network and a prior heatmap. We proposed a relaxed Mean of half channel strategy that allows to efficiently guide the model attribution without limiting its freedom to focus on certain localizations. Through a thorough experimental validation on several challenging FER datasets, we demonstrated that guiding a model attribution with the proposed PAL significantly and consistently improves over existing baselines as well as current state-of-the-art methods, without requiring any additional information at inference time.

This simple method could easily be adapted to different computer vision domains with carefully chosen priors to constrain spatial information within the model. Furthermore, we could imagine working on data where the important spatial information is less easily identifiable at test time, like ultrasound scans for instance, where the crucial findings can be located in different areas for each input image.

\section{Acknowledgement}

This work has been supported by ANR (FacIL, project ANR-17-CE33-0002), by the IUIS institute of Sorbonne Univ. and by the EIT-Health Innovation program (bp2022 \#220648)







\bibliographystyle{IEEEtran}
\bibliography{egbib}
%



\end{document}